# Domain Adaptation Extreme Learning Machines for Drift Compensation in E-nose Systems

Lei Zhang, *Member, IEEE* and David Zhang, *Fellow, IEEE*

*Abstract*—This paper addresses an important issue, known as sensor drift that behaves a nonlinear dynamic property in electronic nose (E-nose), from the viewpoint of machine learning. Traditional methods for drift compensation are laborious and costly due to the frequent acquisition and labeling process for gases samples recalibration. Extreme learning machines (ELMs) have been confirmed to be efficient and effective learning techniques for pattern recognition and regression. However, ELMs primarily focus on the supervised, semi-supervised and unsupervised learning problems in single domain (i.e. source domain). To our best knowledge, ELM with cross-domain learning capability has never been studied. This paper proposes a unified framework, referred to as Domain Adaptation Extreme Learning Machine (DAELM), which learns a robust classifier by leveraging a limited number of labeled data from target domain for drift compensation as well as gases recognition in E-nose systems, without loss of the computational efficiency and learning ability of traditional ELM. In the unified framework, two algorithms called DAELM-S and DAELM-T are proposed for the purpose of this paper, respectively. In order to percept the differences among ELM, DAELM-S and DAELM-T, two *remarks* are provided. Experiments on the popular sensor drift data with multiple batches collected by E-nose system clearly demonstrate that the proposed DAELM significantly outperforms existing drift compensation methods without cumbersome measures, and also bring new perspectives for ELM.

## I. INTRODUCTION

EXTREME learning machine (ELM), proposed for solving a single layer feed-forward network (SLFN) by Huang et al [1, 2], has been proven to be effective and efficient algorithms for pattern classification and regression in different fields. ELM can analytically determine the output weights between the hidden layer and output layer using Moore-Penrose generalized inverse by adopting the square loss of prediction error, which then turns into solving a regularized least square problem efficiently in closed form. The hidden layer output is activated by an infinitely differentiable function with randomly selected input weights and biases of the hidden layer. Huang [3] rigorously proved that the input weights and hidden layer biases can be randomly assigned if the activation function is infinitely differentiable, and also showed that single SLFN with randomly generated additive or RBF nodes with such activation functions can universally approximate any continuous function on any compact subspace of Euclidean space [4].

In recent years, ELM has witnessed a number of improved versions in models, algorithms and real-world applications. ELM shows a comparable or even higher prediction accuracy than that of SVMs which solves a quadratic programming problem. In [3], their differences have been discussed. Some specific examples of improved ELMs have been listed as follows. As the output weights are computed with predefined input weights and biases, a set of non-optimal input weights and hidden biases may exist. Additionally, ELM may require more hidden neurons than conventional learning algorithms in some special applications. Therefore, Zhu et al [5] proposed an evolutionary ELM for more compact networks that speed the response of trained networks. In terms of the imbalanced number of classes, a weighted ELM was proposed for binary/multiclass classification tasks with both balanced and imbalanced data distribution [6]. Due to that the solution of ELM is dense which will require longer time for training in large scale applications, Bai et al [7] proposed a sparse ELM for reducing storage space and testing time. Besides, Li et al [8] also proposed a fast sparse approximation of ELM for sparse classifiers training at a rather low complexity without reducing the generalization performance. For all the versions of ELM mentioned above, supervised learning framework was widely explored in application which limits its ability due to the difficulty in obtaining the labeled data. Therefore, Huang et al [9] proposed a semi-supervised ELM for classification, in which a manifold regularization with graph Laplacian was set, and an unsupervised ELM was also explored for clustering.

In the past years, the contributions to ELM theories and applications have been made substantially by researchers from various fields. However, with the rising of big data, the data distribution obtained in different stages with different experimental conditions may change, i.e. from different domains. It is also well know that E-nose data collection and data labeling is tedious and labor ineffective, while the classifiers trained by a small number of labeled data are not robust and therefore lead to weak generalization, especially for large-scale application. Though ELM performs better generalization when a number of labeled data from source

This work was supported by National Natural Science Foundation of China (61401048), Hong Kong Scholar Program (XJ2013044) and China Postdoctoral Science Foundation (2014M550457).

L. Zhang is with College of Computer Science, Chongqing University, Chongqing 400044, China and also with Department of Computing, The Hong Kong Polytechnic University, Hong Kong (e-mail: leizhang@cqu.edu.cn)

D. Zhang is with Department of Computing, The Hong Kong Polytechnic University, Hong Kong (e-mail: csdzhang@comp.polyu.edu.hk).

domain is used in learning, the transferring capability of ELM is reduced with a limited number of labeled training instances from target domains. Domain adaptation methods have been proposed for robust classifiers learning by leveraging a few labeled instances from target domains [10-14] in machine learning community and computer vision [15]. It is worth noting that domain adaptation is different from semi-supervised learning which assumes that the labeled and unlabeled data are from the same domain in classifier training.

In this paper, we extend ELMs to handle domain adaptation problems for improving the transferring capability of ELM between multiple domains with very few labeled guide instances in target domain, and overcome the generalization disadvantages of ELM in multi-domains application. Specifically, we address the problem of sensor drift compensation in E-nose by using the proposed cross-domain learning framework. Inspired by ELM and knowledge adaptation, a unified domain adaptation ELM framework is proposed for sensor drift compensation. The merits of this paper include:

- To the best of our knowledge, there is no report that couples domain adaptation with ELM framework in machine learning community, while this paper provides several new perspectives for exploring ELM theory.
- We integrate a new methodology i.e. domain adaptation ELM in E-nose for sensor drift compensation and gas recognition. The propose DAELM is a unified classifier learning framework with knowledge adaptability and well addressed the problem of drift as well as gas recognition.
- One method of DAELM, called source domain adaptation ELM (DAELM-S) which learns a classifier by using a number of labeled data from the source domain, and leveraging a limited number of labeled samples from target domain as regularization, is proposed intuitively.
- Another method of DAELM, called target domain adaptation ELM (DAELM-T) is also proposed. DAELM-T learns a classifier using a limited number of labeled instances from target domain, while the remaining numerous unlabeled data are also fully exploited by approximating the prediction of a pre-learned base classifier trained in source domain to that of the learned classifier, into which many existing classifiers can be incorporated as the base classifier.
- Both DAELM-S and DAELM-T can be formed into a unified ELM framework, in which two steps including random feature mapping and output weights training are referred and our DAELM holds the merits of ELM. In both methods, the final solution can be analytically determined, and the generalization performance is guaranteed in E-nose application.

The rest of this paper is organized as follows. In Section II, related work in sensor drift compensation and a brief review of ELM are presented. In Section III, the proposed DAELM framework including two specific algorithms: DAELM-S and DAELM-T is presented. In Section IV, we present the experiments on the popular sensor drift data collected by an E-nose for 3 years, and the results of drift compensation and gas recognition. Finally, Section V concludes the paper.

## II. RELATED WORK

### A. Sensor Drift Compensation in Electronic Nose

Electronic nose is an intelligent multi-sensor system or artificial olfaction system, which is developed as instrument for gas recognition [18, 19], tea quality assessment [20, 21], medical diagnosis [22], environmental monitor and gas concentration estimation [23, 24], etc. by coupling with pattern recognition and gas sensor array with cross-sensitivity and broad spectrum characteristics. An excellent overview of the E-nose and techniques for processing the sensor responses can be referred to as [32], [33].

However, sensors are often operated over a long period in real-world application and lead to aging that seriously reduces the lifetime of sensors. This is so called sensor drift caused by unknown dynamic process, such as poisoning, aging or environmental variations [34]. Sensor drift has deteriorated the performance of classifiers [25] used for gas recognition of chemosensory systems or E-noses, and plagued the sensory community for many years. Therefore, researchers have to re-train the classifier using a number of new samples in a period regularly for recalibration. However, the tedious work for classifier retraining and acquisition of new labeled samples regularly seems to be impossible for recalibration, due to the complicated gaseous experiments of E-nose and labor cost.

The drift problem can be formulated as follows.

Suppose $\mathcal{D}_1, \mathcal{D}_2, \cdots, \mathcal{D}_K$ are gas sensor data sets collected by an E-nose with $K$ batches ranked according to the time intervals, where $\mathcal{D}_i = \{\mathbf{x}_j^i\}_{j=1}^{N_i}, i = 1, \cdots, K$, $\mathbf{x}_j^i$ denotes a feature vector of the $j$-th sample in batch $i$, and $N_i$ is the number of samples in batch $i$. The sensor drift problem is that the feature distributions of $\mathcal{D}_2, \cdots, \mathcal{D}_K$ do not obey the distribution of $\mathcal{D}_1$. As a result, the classifier trained using the labeled data of $\mathcal{D}_1$ has degraded performance when tested on $\mathcal{D}_2, \cdots, \mathcal{D}_K$ due to the deteriorated generalization ability caused by drift. Generally, the mismatch of distribution between $\mathcal{D}_1$ and $\mathcal{D}_i$ becomes larger with increasing batch index $i$ ($i>1$) and aging. From the angle of domain adaptation, in this paper, $\mathcal{D}_1$ is called source domain/auxiliary domain (without drift) with labeled data, $\mathcal{D}_2, \cdots, \mathcal{D}_K$ are referred to as target domain (drifted) in which only a limited number of labeled data is available.

Drift compensation has been studied for many years. Generally, drift compensation methods can be divided into three categories: component correction methods, adaptive methods, and machine learning methods. Specifically, multivariate component correction, such as CCPCA [35] which attempts to find the drift direction using PCA and remove the drift component is recognized as a popular method in periodic calibration. However, CCPCA assumes that the data from all classes behaves in the same way in the presence of drift that is not always the case. Additionally, evolutionary algorithm which optimizes a multiplicative correction factor for drift compensation [36] was proposed as an adaptive method. However, the generalization performance of the correction factor is limited for on-line use due to the nonlinear dynamic behavior of sensor drift. Classifier ensemble in machine learning was first proposed in [26] for drift compensation, which has shown improved gas recognition accuracy using the data with long term drift. An overview of the drift compensation is referred to as [25]. Other recent methods to cope with drift can be referred to as [27]-[29].

Though researchers have paid more attention to sensor drift and aim to find some measures for drift compensation, sensor drift is still a challenging issue in machine olfaction community and sensory field. To our best knowledge, the existing methods are limited in dealing with sensor drift due to their weak generalization to completely new data in presence of drift. Therefore, we aim to enhance the adaptive performance of classifiers to new drifting/drifted data using cross-domain learning with very low complexity. It would be very meaningful and interesting to train a classifier using very few labeled new samples (target domain) without giving up the recognized "useless" old data (source domain), and realize effective and efficient knowledge transfer (i.e. drift compensation) from source domain to multiple target domains.

*B. Principle of ELM*

Given $N$ samples $[\mathbf{x}_1, \mathbf{x}_2, \cdots, \mathbf{x}_N]$ and their corresponding ground truth $[\mathbf{t}_1, \mathbf{t}_2, \cdots, \mathbf{t}_N]$, where $\mathbf{x}_i = [x_{i1}, x_{i1}, \cdots, x_{in}]^T \in \mathbb{R}^n$ and $\mathbf{t}_i = [t_{i1}, t_{i1}, \cdots, t_{im}]^T \in \mathbb{R}^m$, $n$ and $m$ denote the number of input and output neurons, respectively. The output of the hidden layer is denoted as $\mathcal{H}(\mathbf{x}_i) \in \mathbb{R}^{1 \times L}$, where $L$ is the number of hidden nodes and $\mathcal{H}(\cdot)$ is the activation function (e.g. RBF function, sigmoid function). The output weights between the hidden layer and the output layer being learned is denoted as $\boldsymbol{\beta} \in \mathbb{R}^{L \times m}$.

Regularized ELM aims to solve the output weights by minimizing the squared loss summation of prediction errors and the norm of the output weights for over-fitting control, formulated as follows

$$\begin{cases} \min_{\boldsymbol{\beta}} \mathcal{L}_{ELM} = \frac{1}{2}\|\boldsymbol{\beta}\|^2 + C \cdot \frac{1}{2} \cdot \sum_{i=1}^{N}\|\boldsymbol{\xi}_i\|^2 \\ s.t. \ \mathcal{H}(\mathbf{x}_i)\boldsymbol{\beta} = \mathbf{t}_i - \boldsymbol{\xi}_i, i = 1, \dots, N \end{cases} \quad (1)$$

where $\boldsymbol{\xi}_i$ denotes the prediction error w.r.t. to the $i$-th training pattern, and $C$ is a penalty constant on the training errors.

By substituting the constraint term in (1) into the objective function, an equivalent unconstrained optimization problem can be obtained as follows

$$\min_{\boldsymbol{\beta} \in \mathbb{R}^{L \times m}} \mathcal{L}_{ELM} = \frac{1}{2}\|\boldsymbol{\beta}\|^2 + C \cdot \frac{1}{2} \cdot \|\mathbf{T} - \mathbf{H}\boldsymbol{\beta}\|^2 \quad (2)$$

where $\mathbf{H} = [\mathcal{H}(\mathbf{x}_1); \mathcal{H}(\mathbf{x}_2); \dots; \mathcal{H}(\mathbf{x}_N)] \in \mathbb{R}^{N \times L}$ and $\mathbf{T} = [\mathbf{t}_1, \mathbf{t}_2, \dots, \mathbf{t}_N]^T$.

The optimization problem (2) is a well known regularized least square problem. The closed form solution of $\boldsymbol{\beta}$ can be easily solved by setting the gradient of the objective function (2) with respect to $\boldsymbol{\beta}$ to zero.

There are two cases when solving $\boldsymbol{\beta}$, i.e. if the number $N$ of training patterns is larger than $L$, the gradient equation is over-determined, and the closed form solution can be obtained as

$$\boldsymbol{\beta}^* = \left(\mathbf{H}^T\mathbf{H} + \frac{\mathbf{I}_L}{C}\right)^{-1} \mathbf{H}^T \mathbf{T} \quad (3)$$

where $\mathbf{I}_{L \times L}$ denotes the identity matrix.

If the number $N$ of training patterns is smaller than $L$, an under-determined least square problem would be handled. In this case, the solution of (2) can be obtained as

$$\boldsymbol{\beta}^* = \mathbf{H}^T \left(\mathbf{H}\mathbf{H}^T + \frac{\mathbf{I}_N}{C}\right)^{-1} \mathbf{T} \quad (4)$$

where $\mathbf{I}_{N \times N}$ denotes the identity matrix.

Therefore, in classifier training of ELM, the output weights can be computed by using (3) or (4) depending on the number of training instances and the number of hidden nodes. We refer interested readers to as [1] for more details of ELM theory and the algorithms.

III. PROPOSED DOMAIN ADAPTATION ELM FRAMEWORK

In this section, we present formulation of the proposed domain adaptation ELM framework, in which two methods referred to as Source Domain Adaptation ELM (DAELM-S) and Target Domain Adaptation ELM (DAELM-T) are introduced with their learning algorithms, respectively.

*A. Source Domain Adaptation ELM (DAELM-S)*

Suppose that the source domain and target domain are represented by "*S*" and "*T*". In this paper, we assume that all the samples in the source domain are labeled data.

The proposed DAELM-S aims to learn a classifier $\boldsymbol{\beta}_S$ using all labeled instances from the source domain by leveraging a limited number of labeled data from target domain. The DAELM-S can be formulated as

$$\min_{\boldsymbol{\beta}_S, \boldsymbol{\xi}_S^i, \boldsymbol{\xi}_T^j} \frac{1}{2}\|\boldsymbol{\beta}_S\|^2 + C_S \frac{1}{2} \sum_{i=1}^{N_S}\|\boldsymbol{\xi}_S^i\|^2 + C_T \frac{1}{2} \sum_{j=1}^{N_T}\|\boldsymbol{\xi}_T^j\|^2 \quad (5)$$

$$s.t. \begin{cases} \mathbf{H}_S^i \boldsymbol{\beta}_S = \mathbf{t}_S^i - \boldsymbol{\xi}_S^i, i = 1, \dots, N_S \\ \mathbf{H}_T^j \boldsymbol{\beta}_S = \mathbf{t}_T^j - \boldsymbol{\xi}_T^j, j = 1, \dots, N_T \end{cases} \quad (6)$$

where $\mathbf{H}_S^i \in \mathbb{R}^{1 \times L}, \boldsymbol{\xi}_S^i \in \mathbb{R}^{1 \times m}, \mathbf{t}_S^i \in \mathbb{R}^{1 \times m}$ denote the output of hidden layer, the prediction error and the label w.r.t. the $i$-th training instance $\mathbf{x}_S^i$ from the source domain, $\mathbf{H}_T^j \in \mathbb{R}^{1 \times L}, \boldsymbol{\xi}_T^j \in \mathbb{R}^{1 \times m}, \mathbf{t}_T^j \in \mathbb{R}^{1 \times m}$ denote the output of hidden layer, the prediction error and the label vector with respect to the $j$-th guide samples $\mathbf{x}_T^j$ from the target domain, $\boldsymbol{\beta}_S \in \mathbb{R}^{L \times m}$ is the output weights being solved, $N_S$ and $N_T$ denote the number of training instances and guide samples from the source domain and target domain, respectively, $C_S$ and $C_T$ are the penalty coefficients on the prediction errors of the labeled training data from source domain and target domain, respectively. In this paper, $\mathbf{t}_S^{i,j} = 1$ if pattern $\mathbf{x}_i$ belongs to the $j$-th class, and -1 otherwise. For example, $\mathbf{t}_S^i = [1, -1, \cdots, -1]^T$ if $\mathbf{x}_i$ is belong to class 1.

From (5), we can find that the very few labeled guide samples from target domain can make the learning of $\boldsymbol{\beta}_S$ "transferable" and realize the knowledge transfer between source domain and target domain by introducing the third term as regularization coupling with the second constraint in (6), which makes the feature mapping of the guide samples from target domain approximate the labels recognized with the output weights $\boldsymbol{\beta}_S$. The structure of the proposed DAELM-S algorithm to learn $M$ classifiers is illustrated in Fig.1.

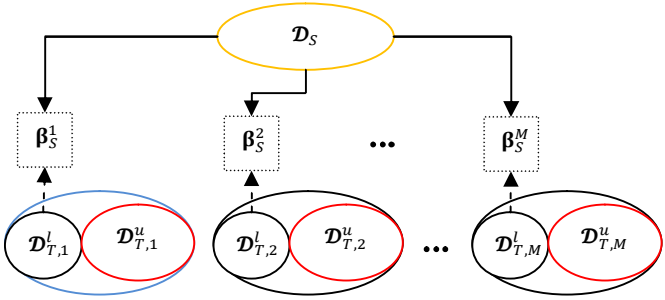

Fig. 1. Structure of DAELM-S algorithm with $M$ target domains ($M$ tasks). The solid arrow denotes the training data from source domain $\mathcal{D}_S$ and the dashed arrow denotes the labeled data from target domain $\mathcal{D}_T^l$ for classifier learning. The unlabeled data from target domain $\mathcal{D}_T^u$ are not used in learning.

---

**Algorithm 1**. DAELM-S
**Input:**
  Training samples $\{\mathbf{X}_S, \mathbf{t}_S\} = \{\mathbf{x}_S^i, \mathbf{t}_S^i\}_{i=1}^{N_S}$ of the source domain $S$;
  Labeled guide samples $\{\mathbf{X}_T, \mathbf{t}_T\} = \{\mathbf{x}_T^j, \mathbf{t}_T^j\}_{j=1}^{N_T}$ of the target domain $T$;
  The tradeoff parameter $C_S$ and $C_T$ for source and target domain.
**Output:**
  The output weights $\boldsymbol{\beta}_S$;
  The predicted output $\mathbf{y}_{Tu}$ of unlabeled data in target domain.
**Procedure:**
1. Initialize the ELM network of $L$ hidden neurons with random input weights $\mathbf{W}$ and hidden bias $\mathbf{B}$.
2. Calculate the output matrix $\mathbf{H}_S$ and $\mathbf{H}_T$ of hidden layer with source and target domains as $\mathbf{H}_S = \mathcal{H}(\mathbf{W} \cdot \mathbf{X}_S + \mathbf{B})$ and $\mathbf{H}_T = \mathcal{H}(\mathbf{W} \cdot \mathbf{X}_T + \mathbf{B})$.
3. **If** $N_S < L$, compute the output weights $\boldsymbol{\beta}_S$ using (12);
   **Else**, compute the output weights $\boldsymbol{\beta}_S$ using (13).
4. Calculate the predicted output $\mathbf{y}_{Tu}$ using (16).
**Return** The output weights $\boldsymbol{\beta}_S$ and predicted output $\mathbf{y}_{Tu}$.

---

To solve the optimization (5), the Largange multiplier equation is formulated as

$$L(\boldsymbol{\beta}_S, \xi_S^i, \xi_T^j, \alpha_S, \alpha_T) = \frac{1}{2}\|\boldsymbol{\beta}_S\|^2 + \frac{C_S}{2}\sum_{i=1}^{N_S}\|\xi_S^i\|^2 + \frac{C_T}{2}\sum_{j=1}^{N_T}\|\xi_T^j\|^2 - \alpha_S(\mathbf{H}_S^i\boldsymbol{\beta}_S - \mathbf{t}_S^i + \xi_S^i) - \alpha_T(\mathbf{H}_T^i\boldsymbol{\beta}_T - \mathbf{t}_T^i + \xi_T^i) \quad (7)$$

where $\alpha_S$ and $\alpha_T$ denote the multiplier vectors.

By setting the partial derivation with respect to $\boldsymbol{\beta}_S, \xi_S^i, \xi_T^j, \alpha_S, \alpha_T$ as zero, we have

$$\begin{cases} \frac{\partial L}{\partial \boldsymbol{\beta}_S} = 0 \rightarrow \boldsymbol{\beta}_S = \mathbf{H}_S^T\alpha_S + \mathbf{H}_T^T\alpha_T \\ \frac{\partial L}{\partial \xi_S} = 0 \rightarrow \alpha_S = C_S\xi_S^T \\ \frac{\partial L}{\partial \xi_T} = 0 \rightarrow \alpha_T = C_T\xi_T^T \\ \frac{\partial L}{\partial \alpha_S} = 0 \rightarrow \mathbf{H}_S\boldsymbol{\beta}_S - \mathbf{t}_S + \xi_S = 0 \\ \frac{\partial L}{\partial \alpha_T} = 0 \rightarrow \mathbf{H}_T\boldsymbol{\beta}_S - \mathbf{t}_T + \xi_T = 0 \end{cases} \quad (8)$$

where $\mathbf{H}_S$ and $\mathbf{H}_T$ are the output matrix of hidden layer with respect to the labeled data from source domain and target domain, respectively. To analytically determine $\boldsymbol{\beta}_S$, the multiplier vectors $\alpha_S$ and $\alpha_T$ should be solved first.

For the case that the number of training samples $N_S$ is smaller than $L$ ($N_S < L$), then $\mathbf{H}_S$ will have more columns than rows and be of full row rank, which leads to a under-determined least square problem, and infinite number of solutions may be obtained. To handle this problem, we substitute the 1st, 2nd, and 3rd equations into the 4th and 5th equations considering that $\mathbf{HH}^T$ is invertible, and then there is

$$\begin{cases} \mathbf{H}_T\mathbf{H}_S^T\alpha_S + \left(\mathbf{H}_T\mathbf{H}_T^T + \frac{\mathbf{I}}{C_T}\right)\alpha_T = \mathbf{t}_T \\ \mathbf{H}_S\mathbf{H}_T^T\alpha_T + \left(\mathbf{H}_S\mathbf{H}_S^T + \frac{\mathbf{I}}{C_S}\right)\alpha_S = \mathbf{t}_S \end{cases} \quad (9)$$

Let $\mathbf{H}_T\mathbf{H}_S^T = \mathcal{A}, \mathbf{H}_T\mathbf{H}_T^T + \frac{\mathbf{I}}{C_T} = \mathcal{B}, \mathbf{H}_S\mathbf{H}_T^T = \mathcal{C}, \mathbf{H}_S\mathbf{H}_S^T + \frac{\mathbf{I}}{C_S} = \mathcal{D}$, then Eq.(9) can be written as

$$\begin{cases} \mathcal{A}\alpha_S + \mathcal{B}\alpha_T = \mathbf{t}_T \\ \mathcal{C}\alpha_T + \mathcal{D}\alpha_S = \mathbf{t}_S \end{cases} \rightarrow \begin{cases} \mathcal{B}^{-1}\mathcal{A}\alpha_S + \alpha_T = \mathcal{A}^{-1}\mathbf{t}_T \\ \mathcal{C}\alpha_T + \mathcal{D}\alpha_S = \mathbf{t}_S \end{cases} \quad (10)$$

Then $\alpha_S$ and $\alpha_T$ can be solved as

$$\begin{cases} \alpha_S = (\mathcal{CB}^{-1}\mathcal{A} - \mathcal{D})^{-1}(\mathcal{CB}^{-1}\mathbf{t}_T - \mathbf{t}_S) \\ \alpha_T = \mathcal{B}^{-1}\mathbf{t}_T - \mathcal{B}^{-1}\mathcal{A}(\mathcal{CB}^{-1}\mathcal{A} - \mathcal{D})^{-1}(\mathcal{CB}^{-1}\mathbf{t}_T - \mathbf{t}_S) \end{cases} \quad (11)$$

Consider the 1st equation in (8), we obtain the output weights as

$$\begin{aligned}\boldsymbol{\beta}_S &= \mathbf{H}_S^T\alpha_S + \mathbf{H}_T^T\alpha_T \\ &= \mathbf{H}_S^T(\mathcal{CB}^{-1}\mathcal{A} - \mathcal{D})^{-1}(\mathcal{CB}^{-1}\mathbf{t}_T - \mathbf{t}_S) + \mathbf{H}_T^T[\mathcal{B}^{-1}\mathbf{t}_T - \mathcal{B}^{-1}\mathcal{A}(\mathcal{CB}^{-1}\mathcal{A} - \mathcal{D})^{-1}(\mathcal{CB}^{-1}\mathbf{t}_T - \mathbf{t}_S)]\end{aligned} \quad (12)$$

where $\mathbf{I}$ is the identity matrix with size of $N_S$.

For the case that the number of training samples $N_S$ is larger than $L$ ($N_S > L$), $\mathbf{H}_S$ has more rows than columns and is of full column rank, which is an over-determined least square problem. Then, we can obtain from the 1st equation in (8) that $\alpha_S = (\mathbf{H}_S\mathbf{H}_S^T)^{-1}(\mathbf{H}_S\boldsymbol{\beta}_S - \mathbf{H}_S\mathbf{H}_T^T\alpha_T)$, after which is substituted into the 4th and 5th equations, we can calculate the output weights $\boldsymbol{\beta}_S$ as follows

$$\begin{cases} \mathbf{H}_S\boldsymbol{\beta}_S + \xi_S = \mathbf{t}_S \\ \mathbf{H}_T\boldsymbol{\beta}_S + \xi_T = \mathbf{t}_T \end{cases} \rightarrow \begin{cases} \mathbf{H}_S^T\mathbf{H}_S\boldsymbol{\beta}_S + \frac{1}{C_S}\mathbf{H}_S^T\alpha_S = \mathbf{H}_S^T\mathbf{t}_S \\ \mathbf{H}_T\boldsymbol{\beta}_S + \frac{1}{C_T}\alpha_T = \mathbf{t}_T \end{cases}$$

$$\rightarrow \begin{cases} \mathbf{H}_S^T\mathbf{H}_S\boldsymbol{\beta}_S + \frac{1}{C_S}\mathbf{H}_S^T(\mathbf{H}_S\mathbf{H}_S^T)^{-1}(\mathbf{H}_S\boldsymbol{\beta}_S - \mathbf{H}_S\mathbf{H}_T^T\alpha_T) = \mathbf{H}_S^T\mathbf{t}_S \\ \alpha_T = C_T(\mathbf{t}_T - \mathbf{H}_T\boldsymbol{\beta}_S) \end{cases}$$

$$\rightarrow \left(\mathbf{H}_S^T\mathbf{H}_S + \frac{\mathbf{I}}{C_S} + \frac{C_T}{C_S}\mathbf{H}_T^T\mathbf{H}_T\right)\boldsymbol{\beta}_S = \mathbf{H}_S^T\mathbf{t}_S + \frac{C_T}{C_S}\mathbf{H}_T^T\mathbf{t}_T$$

$$\rightarrow \boldsymbol{\beta}_S = (\mathbf{I} + C_S\mathbf{H}_S^T\mathbf{H}_S + C_T\mathbf{H}_T^T\mathbf{H}_T)^{-1}(C_S\mathbf{H}_S^T\mathbf{t}_S + C_T\mathbf{H}_T^T\mathbf{t}_T) \quad (13)$$

where $\mathbf{I}$ is the identity matrix with size of $L$.

In fact, the optimization (5) can be reformulated as an equivalent unconstrained optimization problem in matrix form by substituting the constraints into the objective function,

$$\min_{\boldsymbol{\beta}_S} L_{DAELM-S}(\boldsymbol{\beta}_S) = \frac{1}{2}\|\boldsymbol{\beta}_S\|^2 + C_S\frac{1}{2}\|\mathbf{t}_S - \mathbf{H}_S\boldsymbol{\beta}_S\|^2 + C_T\frac{1}{2}\|\mathbf{t}_T - \mathbf{H}_T\boldsymbol{\beta}_S\|^2 \quad (14)$$

By setting the gradient of $L_{DAELM-S}$ w.r.t. $\boldsymbol{\beta}_S$ to be zero,

$$\nabla L_{DAELM-S} = \boldsymbol{\beta}_S - C_S\mathbf{H}_S^T(\mathbf{t}_S - \mathbf{H}_S\boldsymbol{\beta}_S) - C_T\mathbf{H}_T^T(\mathbf{t}_T - \mathbf{H}_T\boldsymbol{\beta}_S) = 0 \quad (15)$$

Then, we can easily solve (15) to obtain $\boldsymbol{\beta}_S$ formulated in (13).

For recognition of the numerous unlabeled data in target domain, we calculate the output of DAELM-S network as

$$\mathbf{y}_{Tu}^k = \mathbf{H}_{Tu}^k \cdot \boldsymbol{\beta}_S, k = 1, \ldots, N_{Tu} \quad (16)$$

where $\mathbf{H}_{Tu}^k$ denote the hidden layer output with respect to the $k$-th unlabeled vector in target domain, and $N_{Tu}$ is the number of unlabeled vectors in target domain. The index corresponding to the maximum value in $\mathbf{y}_{Tu}^k$ is the class of the $k$-th sample.

For implementation, the DAELM-S algorithm is summarized as Algorithm 1.

*B. Target Domain Adaptation ELM (DAELM-T)*

In the proposed DAELM-S, the classifier $\boldsymbol{\beta}_S$ is learned on the source domain with the very few labeled guide samples from the target domain as regularization. However, the unlabeled data is neglected which can also improve the performance of classification [17]. Different from DAELM-S, DAELM-T aims to learn a classifier $\boldsymbol{\beta}_T$ on a very limited number of labeled samples from target domain, by leveraging numerous unlabeled data in target domain, into which a base classifier $\boldsymbol{\beta}_B$ trained by source data is incorporated. The proposed DAELM-T is formulated as

$$\min_{\boldsymbol{\beta}_T} L_{DAELM-T}(\boldsymbol{\beta}_T) = \frac{1}{2}\|\boldsymbol{\beta}_T\|^2 + C_T \frac{1}{2}\|\mathbf{t}_T - \mathbf{H}_T \boldsymbol{\beta}_T\|^2 + C_{Tu} \frac{1}{2}\|\mathbf{H}_{Tu}\boldsymbol{\beta}_B - \mathbf{H}_{Tu}\boldsymbol{\beta}_T\|^2 \quad (17)$$

where $\boldsymbol{\beta}_T$ denotes the learned classifier, $C_T, \mathbf{H}_T, \mathbf{t}_T$ are the same as that in DAELM-S, $C_{Tu}, \mathbf{H}_{Tu}$ denote the regularization parameter and the output matrix of the hidden layer with respect to the unlabeled data in target domain. The first term is to against the over-fitting, the second term is the least square loss function, and the third term is the regularization which means the domain adaptation between source domain and target domain. Note that $\boldsymbol{\beta}_B$ is a base classifier learned with source data. In this paper, regularized ELM is used to train a base classifier $\boldsymbol{\beta}_B$ by solving

$$\min_{\boldsymbol{\beta}_B} L_{ELM}(\boldsymbol{\beta}_B) = \frac{1}{2}\|\boldsymbol{\beta}_B\|^2 + C_S \frac{1}{2}\|\mathbf{t}_S - \mathbf{H}_S \boldsymbol{\beta}_B\|^2 \quad (18)$$

where $C_S, \mathbf{t}_S, \mathbf{H}_S$ denote the same meaning as that in DAELM-S.

The structure of the proposed DAELM-T is described in Fig. 2, from which we can see that the unlabeled data in target domain have also been exploited.

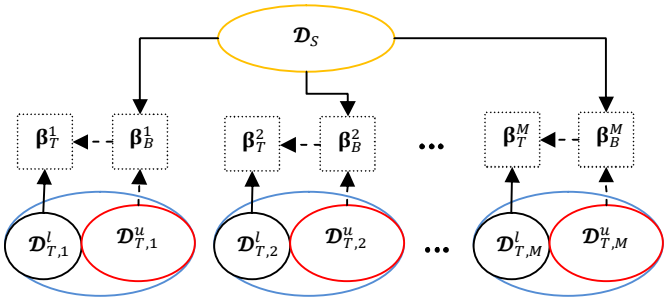

Fig. 2. Structure of DAELM-T algorithm with $M$ target domains ($M$ tasks). The solid arrow connected with $\mathcal{D}_S$ denotes the training for base classifier $\boldsymbol{\beta}_B$, the dashed line connected with $\mathcal{D}_T^u$ denotes the tentative test of base classifier using the unlabeled data from target domain, the solid arrow connected with $\mathcal{D}_T^l$ denotes the terminal classifier learning of $\boldsymbol{\beta}_T$, and the dashed arrow connected between $\boldsymbol{\beta}_B$ and $\boldsymbol{\beta}_T$ denotes the regularization for learning $\boldsymbol{\beta}_T$.

To solve the optimization (17), by setting the gradient of $L_{DAELM-T}$ with respect to $\boldsymbol{\beta}_T$ to be zero, we then have

$$\nabla L_{DAELM-T} =$$

$$\boldsymbol{\beta}_T - C_T \mathbf{H}_T^{\mathrm{T}}(\mathbf{t}_T - \mathbf{H}_T \boldsymbol{\beta}_T) - C_{Tu}\mathbf{H}_{Tu}^{\mathrm{T}}(\mathbf{H}_{Tu}\boldsymbol{\beta}_B - \mathbf{H}_{Tu}\boldsymbol{\beta}_T) = 0 \quad (19)$$

If $N_T > L$, then we can have from (19) that

$$\boldsymbol{\beta}_T = (\mathbf{I} + C_T \mathbf{H}_T^{\mathrm{T}}\mathbf{H}_T + C_{Tu}\mathbf{H}_{Tu}^{\mathrm{T}}\mathbf{H}_{Tu})^{-1}(C_T \mathbf{H}_T^{\mathrm{T}}\mathbf{t}_T + C_{Tu}\mathbf{H}_{Tu}^{\mathrm{T}}\mathbf{H}_{Tu}\boldsymbol{\beta}_B) \quad (20)$$

where $I$ is the identity matrix with size of $L$.

If $N_T < L$, we would like to obtain $\boldsymbol{\beta}_T$ of the proposed DAELM-T according to the solving manner in DAELM-S. Let $\mathbf{t}_{Tu} = \mathbf{H}_{Tu}\boldsymbol{\beta}_B$, the model (17) can be re-written as

$$\min_{\boldsymbol{\beta}_T, \xi_T^i, \xi_{Tu}^j} \frac{1}{2}\|\boldsymbol{\beta}_T\|^2 + C_T \frac{1}{2}\sum_{i=1}^{N_T}\|\xi_T^i\|^2 + C_{Tu}\frac{1}{2}\sum_{j=1}^{N_{Tu}}\|\xi_{Tu}^j\|^2$$
$$\text{s.t.} \begin{cases} \mathbf{H}_T^i \boldsymbol{\beta}_T = \mathbf{t}_T^i - \xi_T^i, i = 1, \ldots, N_T \\ \mathbf{H}_{Tu}^j \boldsymbol{\beta}_T = \mathbf{t}_{Tu}^j - \xi_{Tu}^j, j = 1, \ldots, N_{Tu} \end{cases} \quad (21)$$

The Lagrange multiplier equation of (21) can be written as

$$L(\boldsymbol{\beta}_T, \xi_T^i, \xi_{Tu}^i, \boldsymbol{\alpha}_T, \boldsymbol{\alpha}_{Tu}) = \frac{1}{2}\|\boldsymbol{\beta}_T\|^2 + \frac{C_T}{2}\sum_{i=1}^{N_T}\|\xi_T^i\|^2 + \frac{C_{Tu}}{2}\sum_{j=1}^{N_{Tu}}\|\xi_{Tu}^j\|^2 - \boldsymbol{\alpha}_T(\mathbf{H}_T^i\boldsymbol{\beta}_T - \mathbf{t}_T^i + \xi_T^i) - \boldsymbol{\alpha}_{Tu}(\mathbf{H}_{Tu}^i\boldsymbol{\beta}_T - \mathbf{t}_{Tu}^i + \xi_{Tu}^i) \quad (22)$$

By setting the partial derivation with respect to $\boldsymbol{\beta}_T, \xi_T^i, \xi_{Tu}^j, \boldsymbol{\alpha}_T, \boldsymbol{\alpha}_{Tu}$ to be zero, we have

$$\begin{cases} \frac{\partial L}{\partial \boldsymbol{\beta}_T} = 0 \rightarrow \boldsymbol{\beta}_T = \mathbf{H}_T^{\mathrm{T}}\boldsymbol{\alpha}_T + \mathbf{H}_{Tu}^{\mathrm{T}}\boldsymbol{\alpha}_{Tu} \\ \frac{\partial L}{\partial \xi_T} = 0 \rightarrow \boldsymbol{\alpha}_T = C_T \xi_T^{\mathrm{T}} \\ \frac{\partial L}{\partial \xi_{Tu}} = 0 \rightarrow \boldsymbol{\alpha}_{Tu} = C_{Tu}\xi_{Tu}^{\mathrm{T}} \\ \frac{\partial L}{\partial \boldsymbol{\alpha}_T} = 0 \rightarrow \mathbf{H}_T \boldsymbol{\beta}_T - \mathbf{t}_T + \xi_T = 0 \\ \frac{\partial L}{\partial \boldsymbol{\alpha}_{Tu}} = 0 \rightarrow \mathbf{H}_{Tu}\boldsymbol{\beta}_T - \mathbf{t}_{Tu} + \xi_{Tu} = 0 \end{cases} \quad (23)$$

To solve $\boldsymbol{\beta}_T$, let $\mathbf{H}_{Tu}\mathbf{H}_T^{\mathrm{T}} = \mathcal{O}$, $\mathbf{H}_{Tu}\mathbf{H}_{Tu}^{\mathrm{T}} + \frac{\mathbf{I}}{C_{Tu}} = \mathcal{P}$, $\mathbf{H}_T\mathbf{H}_{Tu}^{\mathrm{T}} = \mathcal{Q}$, and $\mathbf{H}_T\mathbf{H}_T^{\mathrm{T}} + \frac{\mathbf{I}}{C_T} = \mathcal{R}$,

By calculating in the same way as (9), (10), and (11), we get

$$\begin{cases} \boldsymbol{\alpha}_T = (\mathcal{Q}\mathcal{P}^{-1}\mathcal{O} - \mathcal{R})^{-1}(\mathcal{Q}\mathcal{P}^{-1}\mathbf{t}_{Tu} - \mathbf{t}_T) \\ \boldsymbol{\alpha}_{Tu} = \mathcal{P}^{-1}\mathbf{t}_{Tu} - \mathcal{P}^{-1}\mathcal{O}(\mathcal{Q}\mathcal{P}^{-1}\mathcal{O} - \mathcal{R})^{-1}(\mathcal{Q}\mathcal{P}^{-1}\mathbf{t}_{Tu} - \mathbf{t}_T) \end{cases} \quad (24)$$

Therefore, when $N_T < L$, the output weights can be obtained as

$$\boldsymbol{\beta}_T = \mathbf{H}_T^{\mathrm{T}}\boldsymbol{\alpha}_T + \mathbf{H}_{Tu}^{\mathrm{T}}\boldsymbol{\alpha}_{Tu}$$
$$= \mathbf{H}_T^{\mathrm{T}}(\mathcal{Q}\mathcal{P}^{-1}\mathcal{O} - \mathcal{R})^{-1}(\mathcal{Q}\mathcal{P}^{-1}\mathbf{t}_{Tu} - \mathbf{t}_T) + \mathbf{H}_{Tu}^{\mathrm{T}}[\mathcal{P}^{-1}\mathbf{t}_{Tu} - \mathcal{P}^{-1}\mathcal{O}(\mathcal{Q}\mathcal{P}^{-1}\mathcal{O} - \mathcal{R})^{-1}(\mathcal{Q}\mathcal{P}^{-1}\mathbf{t}_{Tu} - \mathbf{t}_T)] \quad (25)$$

where $\mathbf{t}_{Tu} = \mathbf{H}_{Tu}\boldsymbol{\beta}_B$, and $\mathbf{I}$ is the identity matrix with size of $N_T$.

For recognition of the numerous unlabeled data in target domain, we calculate the final output of DAELM-T as

$$\mathbf{y}_{Tu}^k = \mathbf{H}_{Tu}^k \cdot \boldsymbol{\beta}_T, k = 1, \ldots, N_{Tu} \quad (26)$$

where $\mathbf{H}_{Tu}^k$ denote the hidden layer output with respect to the $k$-th unlabeled sample vector in target domain, and $N_{Tu}$ is the number of unlabeled vectors in target domain.

For implementation in experiment, the DAELM-T algorithm is summarized as Algorithm 2.

**Algorithm 2**. DAELM-T

**Input:**
Training samples $\{\mathbf{X}_S, \mathbf{t}_S\} = \{\mathbf{x}_S^i, \mathbf{t}_S^i\}_{i=1}^{N_S}$ of the source domain $S$;
Labeled guide samples $\{\mathbf{X}_T, \mathbf{t}_T\} = \{\mathbf{x}_T^j, \mathbf{t}_T^j\}_{j=1}^{N_T}$ of the target domain $T$;
Unlabeled samples $\{\mathbf{X}_{Tu}\} = \{\mathbf{x}_{Tu}^k\}_{k=1}^{N_{Tu}}$ of the target domain $T$;
The tradeoff parameters $C_S$, $C_T$ and $C_{Tu}$.
**Output:**
The output weights $\boldsymbol{\beta}_T$;
The predicted output $\mathbf{y}_{Tu}$ of unlabeled data in target domain.
**Procedure:**
1. Initialize the ELM network of $L$ hidden neurons with random input weights $\mathbf{W}_1$ and hidden bias $\mathbf{B}_1$.
2. Calculate the output matrix $\mathbf{H}_S$ of hidden layer with source domain as $\mathbf{H}_S = \mathcal{H}(\mathbf{W}_1 \cdot \mathbf{X}_S + \mathbf{B}_1)$.
3. **If** $N_S < L$, compute the output weights $\boldsymbol{\beta}_B$ of the base classifier using (4);
  **Else**, compute the output weights $\boldsymbol{\beta}_B$ of the base classifier using (3).
4. Initialize the ELM network of $L$ hidden neurons with random input weights $\mathbf{W}_2$ and hidden bias $\mathbf{B}_2$.
5. Calculate the hidden layer output matrix $\mathbf{H}_T$ and $\mathbf{H}_{Tu}$ of labeled and unlabeled data in target domains as $\mathbf{H}_T = \mathcal{H}(\mathbf{W}_2 \cdot \mathbf{X}_T + \mathbf{B}_2)$ and $\mathbf{H}_{Tu} = \mathcal{H}(\mathbf{W}_2 \cdot \mathbf{X}_{Tu} + \mathbf{B}_2)$.
6. **If** $N_T < L$, compute the output weights $\boldsymbol{\beta}_T$ using (25);
  **Else**, compute the output weights $\boldsymbol{\beta}_T$ using (20).
7. Calculate the predicted output $\mathbf{y}_{Tu}$ using (26).
**Return** The output weights $\boldsymbol{\beta}_T$ and predicted output $\mathbf{y}_{Tu}$.

*Remark 1:* From the algorithms of DAELM-S and DAELM-T, we observe that the same two stages as ELM are included: (1) feature mapping with randomly selected weights and biases; (2) output weights computation. For ELM, the algorithm is constructed and implemented in a single domain (source domain), as a result, the generalization performance is degraded in new domains. In the proposed DAELM framework, a limited number of labeled samples and numerous unlabeled data in target domain are exploited without changing the unified ELM framework, and the merits of ELM are inherited. The framework for DAELM might draw some new perspectives of domain adaptation for developing ELM theory.

*Remark 2*: We observe that the DAELM-S has similar structure in model and algorithm with DAELM-T. The essential difference lies in that numerous unlabeled data which may be useful for improving generalization performance are exploited in DAELM-T through a pre-learned base classifier. Specifically, DAELM-S learns a classifier using the labeled training data in source domain but draw some new knowledge by leveraging a limited number of labeled samples from target domain, such that the knowledge from target domain can be effectively transferred to source domain for generalization. Whilst DAELM-T attempts to train a classifier using a limited number of labeled data from target domain as "main knowledge" but introduces a regularizer that minimizes the error between outputs of DAELM-T classifier $\boldsymbol{\beta}_T$ and the base classifier $\boldsymbol{\beta}_B$ computed on the unlabeled input data.

## IV. EXPERIMENTS

In this section, we will employ the sensor drift compensation experiment on the E-nose olfactory data by using the proposed DAELM-S and DAELM-T algorithms.

### A. Description of Experimental Data

For verification of the proposed DAELM-S and DAELM-T algorithms, the long-term sensor drift big data of three years that was released in UCI Machine Learning Repository [31] by Vergara *et al.* [26, 30] is exploited and studied in this paper.

The sensor drift big dataset was gathered during the period from January 2008 to February 2011 with 36 months in a gas delivery platform. Totally, this dataset contains 13,910 measurements (observations) from an electronic nose system with 16 gas sensors exposed to 6 kinds of pure gaseous substances including acetone, acetaldehyde, ethanol, ethylene, ammonia, and toluene at different concentration levels, individually. For each sensor, 8 features were extracted, and a 128-dimensional feature vector (8 features × 16 sensors) for each observation is formulated as a result. We refer readers to as [26] for specific technical details on how to select the 8 features for each sensor. In total, 10 batches of sensor data that collected in different time intervals are included in the dataset. The details of the dataset are presented in Table I.

TABLE I
EXPERIMENTAL DATA OF SENSOR DRIFT IN ELECTRONIC NOSE

| Batch ID | Month | Acetone | Acetaldehyde | Ethanol | Ethylene | Ammonia | Toluene | Total |
|---|---|---|---|---|---|---|---|---|
| Batch 1 | 1, 2 | 90 | 98 | 83 | 30 | 70 | 74 | 445 |
| Batch 2 | 3~10 | 164 | 334 | 100 | 109 | 532 | 5 | 1244 |
| Batch 3 | 11, 12, 13 | 365 | 490 | 216 | 240 | 275 | 0 | 1586 |
| Batch 4 | 14, 15 | 64 | 43 | 12 | 30 | 12 | 0 | 161 |
| Batch 5 | 16 | 28 | 40 | 20 | 46 | 63 | 0 | 197 |
| Batch 6 | 17, 18, 19, 20 | 514 | 574 | 110 | 29 | 606 | 467 | 2300 |
| Batch 7 | 21 | 649 | 662 | 360 | 744 | 630 | 568 | 3613 |
| Batch 8 | 22, 23 | 30 | 30 | 40 | 33 | 143 | 18 | 294 |
| Batch 9 | 24, 30 | 61 | 55 | 100 | 75 | 78 | 101 | 470 |
| Batch 10 | 36 | 600 | 600 | 600 | 600 | 600 | 600 | 3600 |

For visualization of the drift behavior existing in the dataset, we first plot the sensor response before and after drifting. We view the data in batch 1 as non-drift, and select batch 2, batch 7, and batch 10 as drifted data, respectively, and the response is given in Fig. 3. It's known that sensor drift shows nonlinear behavior in a multi-dimensional sensor array, and it is impossible to intuitively and directly calibrate the sensor response using some linear or nonlinear transformation. Instead, we consider it as a space distribution adaptation using transfer learning and realize the drift compensation in decision level. Therefore, to observe the space distribution variation with drift, we apply principal component analysis (PCA) on the dataset, and project the data into a 2D subspace based on the first two PCs. The projected 2D subspace for all data in each batch is shown in Fig. 4, from which we can observe the significant changes of data space distribution caused by drift over time.

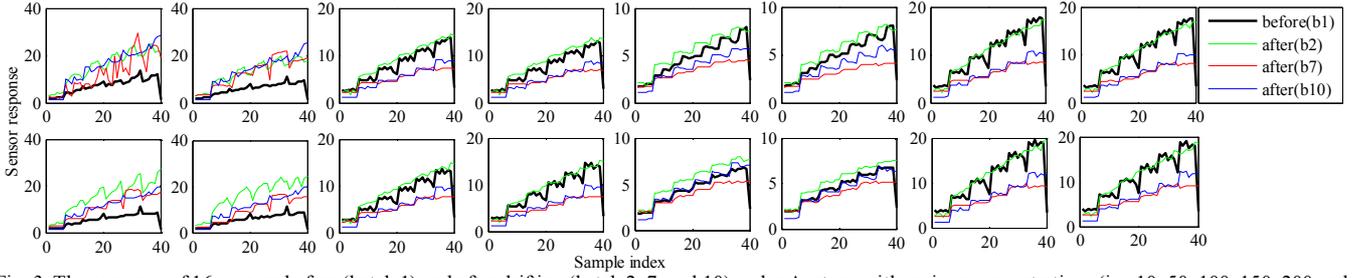

Fig. 3. The response of 16 sensors before (batch 1) and after drifting (batch 2, 7, and 10) under Acetone with various concentrations (i.e. 10, 50, 100, 150, 200 and 250ppm). Totally, 40 samples including 6, 7, 7, 6, 7, and 7 samples for each concentration, respectively, are illustrated for visually shown the drift behavior.

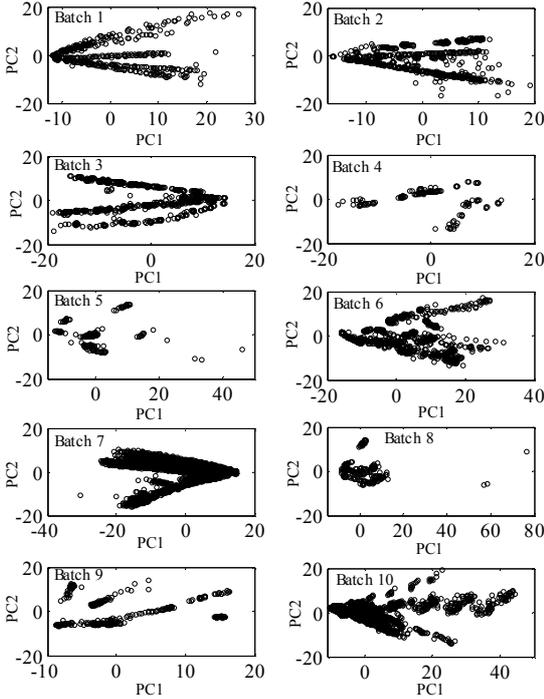

Fig. 4. Principal component space of ten batches for all data. The drift is clearly demonstrated by the different data distribution among batches.

It's worth noting that sensor responses after drift cannot be calibrated directly due to the nonlinear dynamic behavior or chaotic behavior [28] of sensor drift. Therefore, drift compensation in decision level by data distribution adaptation and machine learning is more appealing.

Considering that a small number of labeled samples (guide samples) should be first selected from the target domains in the proposed DAELM-S and DAELM-T algorithms, while the labeled target data plays an important role in knowledge adaptation, we therefore adopt a representative labeled sample selection algorithm (SSA) based on the Euclidean distance $d(x_p, x_q)$ of a sample pair $(x_p, x_q)$. For detail, the SSA algorithm is summarized as Algorithm 3.

The visual SSA algorithm in 2-dimensional coordinate plane for selecting 5 guide samples from each target domain (batch) is shown in Fig. 5 as an example. The patterns marked as "1" denote the first two selected patterns (farthest distance) in Step2. Then, the patterns marked as "2", "3", "4" denote the three selected patterns sequentially. The SSA is for the purpose that the labeled samples selected from target domains should be representative and global in the data space, and promise the generalization performance of domain adaptation.

**Algorithm 3**: SSA algorithm
**Input:**
The data $\mathbf{X}_T$ from target domain;
The predefined number $k$ of labeled samples being selected.
**Output:**
The selected $k$ labeled guide set $\mathcal{S}^l = \{s_1^l, s_2^l, \cdots, s_k^l\}$.
**Procedure:**
**While** the number of selected labeled instances does not reach $k$ **do**
*Step1:* Calculate the Euclidean distance in pair-wise from each target domain, and select the farthest two patterns $s_1^l$ and $s_2^l$ as the labeled instances which is put into the guide set $\mathcal{S}^l = \{s_1^l, s_2^l\}$;
*Step2:* To a pattern $x_i$, calculate the Euclidean distances $d(x_i, \mathcal{S}^l)$, and the nearest distance in each pair for pattern $x_i$ is put into the set $\mathcal{N}_d(x_i)$;
*Step3:* The pattern with the farthest distance in set $\mathcal{N}_d(x_i)$ is then selected as labeled sample $s_3^l$, and update selected labeled guide set $\mathcal{S}^l = \{s_1^l, s_2^l, s_3^l\}$;
*Step4:* **If** the size of the guide set $\mathcal{S}^l$ reaches the number $k$, **break**;
**end while**
**Return** $\mathcal{S}^l = \{s_1^l, s_2^l, \cdots, s_k^l\}$

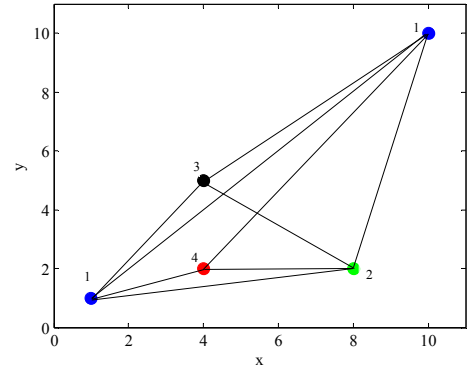

Fig. 5. Visual description of SSA Algorithm 3.

### B. Experimental Setup

We strictly follow the experimental setup in [26] to evaluate our DAELM framework. In default, the number of hidden neurons $L$ is set as 1000, and The RBF function (i.e. *radbas*) with kernel width set as 1 is used as activation function (i.e. feature mapping function) in the hidden layer. The features are scaled appropriately to lie in interval (-1,1). In DAELM-S algorithm, the penalty coefficients $C_S$ and $C_T$ are empirically set as 0.01 and 10 throughout the experiments, respectively. In DAELM-T algorithm, the penalty coefficient $C_S$ for base classifier is set as 0.001, $C_T$ and $C_{Tu}$ are set as 0.001 and 100 throughout the experiments, respectively. For effective verification of the proposed methods, two experimental settings according to [16] are given as follows.

- **Setting-1:** Take batch 1 (source domain) as fixed training set and tested on batch $K$, $K$=2,…,10 (target domains);
- **Setting-2:** The training set (source domain) is

dynamically changed with batch $K$-1 and tested on batch $K$ (target domain), $K$=2,…,10.

Following the two settings, we realize our proposed DAELM framework and compare with multi-class SVM with RBF kernel (SVM-rbf), the geodesic flow kernel (SVM-gfk), and the combination kernel (SVM-comgfk). Besides, we also compared with the semi-supervised methods such as manifold regularization with RBF kernel (ML-rbf) and manifold regularization with combination kernel (ML-comgfk). The above machine learning based methods have been reported for drift compensation [16] using the same dataset. The formulation of geodesic flow kernel as a domain adaptation method can be referred to as [37]. Additionally, the regularized ELM with RBF function in hidden layer (ELM-rbf) from [29] is also compared as baseline in experiments. The popular CC-PCA method [35] and classifier ensemble [26] for drift compensation are also reported in Setting 1 and Setting 2.

Due to the random selection of input weights between input layer and hidden layer, and bias in hidden layer under ELM framework, in experiments, we run the ELM, DAELM-S and DAELM-T for 10 times, and the average values are reported. Note that ELM is trained using the same labeled source data and target data as the proposed DAELM.

TABLE II
COMPARISONS OF RECOGNITION ACCURACY (%) UNDER THE EXPERIMENTAL SETTING 1

| Batch ID | Batch 2 | Batch 3 | Batch 4 | Batch 5 | Batch 6 | Batch 7 | Batch 8 | Batch 9 | Batch 10 | Average |
|---|---|---|---|---|---|---|---|---|---|---|
| CC-PCA | 67.00 | 48.50 | 41.00 | 35.50 | 55.00 | 31.00 | 56.50 | 46.50 | 30.50 | 45.72 |
| SVM-rbf | 74.36 | 61.03 | 50.93 | 18.27 | 28.26 | 28.81 | 20.07 | 34.26 | 34.47 | 38.94 |
| SVM-gfk | 72.75 | 70.08 | 60.75 | 75.08 | 73.82 | 54.53 | 55.44 | 69.62 | 41.78 | 63.76 |
| SVM-comgfk | 74.47 | 70.15 | 59.78 | 75.09 | 73.99 | 54.59 | 55.88 | 70.23 | 41.85 | 64.00 |
| ML-rbf | 42.25 | 73.69 | 75.53 | 66.75 | 77.51 | 54.43 | 33.50 | 23.57 | 34.92 | 53.57 |
| ML-comgfk | 80.25 | 74.99 | 78.79 | 67.41 | 77.82 | 71.68 | 49.96 | 50.79 | 53.79 | 67.28 |
| ELM-rbf | 70.63 | 66.44 | 66.83 | 63.45 | 69.73 | 51.23 | 49.76 | 49.83 | 33.50 | 57.93 |
| **Our DAELM-S(20)** | 87.57 | 96.53 | 82.61 | 81.47 | 84.97 | 71.89 | 78.10 | 87.02 | 57.42 | **80.84** |
| **Our DAELM-S(30)** | 87.98 | 95.74 | 85.16 | 95.99 | 94.14 | 83.51 | 86.90 | 100.0 | 53.62 | **87.00** |
| **Our DAELM-T(40)** | 83.52 | 96.34 | 88.20 | 99.49 | 78.43 | 80.93 | 87.42 | 100.0 | 56.25 | **85.62** |
| **Our DAELM-T(50)** | 97.96 | 95.34 | 99.32 | 99.24 | 97.03 | 83.09 | 95.27 | 100.0 | 59.45 | **91.86** |

TABLE III
COMPARISONS OF RECOGNITION ACCURACY (%) UNDER THE EXPERIMENTAL SETTING 2

| Batch ID | 1→2 | 2→3 | 3→4 | 4→5 | 5→6 | 6→7 | 7→8 | 8→9 | 9→10 | Average |
|---|---|---|---|---|---|---|---|---|---|---|
| SVM-rbf | 74.36 | 87.83 | 90.06 | 56.35 | 42.52 | 83.53 | 91.84 | 62.98 | 22.64 | 68.01 |
| SVM-gfk | 72.75 | 74.02 | 77.83 | 63.91 | 70.31 | 77.59 | 78.57 | 86.23 | 15.76 | 68.56 |
| SVM-comgfk | 74.47 | 73.75 | 78.51 | 64.26 | 69.97 | 77.69 | 82.69 | 85.53 | 17.76 | 69.40 |
| ML-rbf | 42.25 | 58.51 | 75.78 | 29.10 | 53.22 | 69.17 | 55.10 | 37.94 | 12.44 | 48.17 |
| ML-comgfk | 80.25 | 98.55 | 84.89 | 89.85 | 75.53 | 91.17 | 61.22 | 95.53 | 39.56 | 79.62 |
| Ensemble | 74.40 | 88.00 | 92.50 | 94.00 | 69.00 | 69.50 | 91.00 | 77.00 | 65.00 | 80.04 |
| ELM-rbf | 70.63 | 40.44 | 64.16 | 64.37 | 72.70 | 80.75 | 88.20 | 67.00 | 22.00 | 63.36 |
| **Our DAELM-S(20)** | 87.57 | 96.90 | 85.59 | 95.89 | 80.53 | 91.56 | 88.71 | 88.40 | 45.61 | **84.53** |
| **Our DAELM-S(30)** | 87.98 | 96.58 | 89.75 | 99.04 | 84.43 | 91.75 | 89.83 | 100.0 | 58.44 | **88.64** |
| **Our DAELM-T(40)** | 83.52 | 96.41 | 81.36 | 96.45 | 85.13 | 80.49 | 85.71 | 100.0 | 56.81 | **85.10** |
| **Our DAELM-T(50)** | 97.96 | 95.62 | 99.63 | 98.17 | 97.13 | 83.10 | 94.90 | 100.0 | 59.88 | **91.82** |

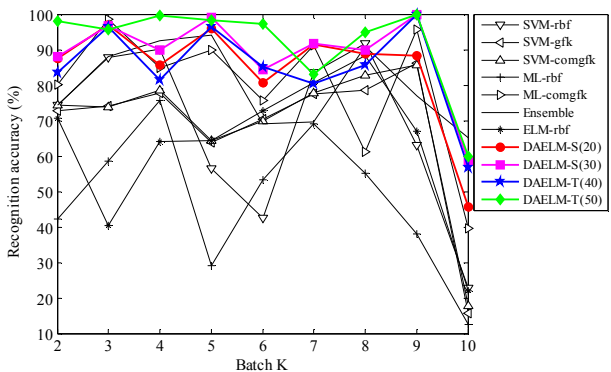

Fig. 6. Comparisons of different methods in experimental **Setting 1**.

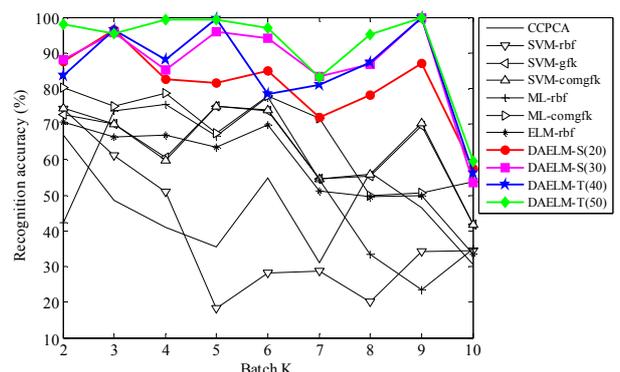

Fig. 7. Comparisons of different methods in experimental **Setting 2.**

*C. Results and Comparisons*

We conduct the experiments and discussion on **Setting 1** and **Setting 2**, respectively. The recognition results of 9 batches for different methods under experimental setting 1 are reported in Table II. We consider two conditions of DAELM-S with 20 labeled target samples and 30 labeled target samples, respectively. For DAELM-T, 40 and 50 labeled samples from the target domain are used, respectively, considering that DAELM-T trains a classifier only using a limited number of labeled samples from target domain. For visually observing the performance of all methods, we show the recognition accuracy on batches successively as Fig.6. From Table II and Fig.6, we have the following observations:

1. SVM with the combined kernel of geodesic flow kernels (SVM-comgfk) performs better results than the popular CC-PCA method and other SVM based methods in most batches, except the results of batch 4 and batch 8. It demonstrates that machine learning methods show more usefulness in drift compensation than traditional calibration.
2. Manifold learning with combined kernel (ML-comgfk) obtains an average accuracy of 67.3% and outperforms all baseline methods. It demonstrates that manifold regularization and combined kernel are more effective in semi-supervised learning with a limited number of samples.
3. The generalization performance and knowledge transfer capability of regularized ELM have been well improved by the proposed DAELM. The results of our DAELM-S and DAELM-T have an average improvement of about 30% in recognition accuracy than traditional ELM. The highest recognition accuracy of 91.86% under sensor drift is obtained using our proposed algorithm.
4. Both the proposed DAELM-S and DAELM-T significantly outperform all other existing methods including traditional CC-PCAM, SVM and manifold regularization based machine learning methods. In addition, DAELM-T(50) has an obvious improvement than DAELM-T(40) and DAELM-S, which shows that more labeled target data is expected for DAELM-T. While DAELM-S can also perform well comparatively with fewer labeled target data. From the computations, due to that a base classifier is first trained in DAELM-T and more labeled target data need, DAELM-S maybe a better choice in realistic applications.

From the experimental results in experimental **Setting 1**, the proposed methods outperform all other methods in drift compensation. We then follow the experimental **Setting 2**, i.e. trained on batch $K$-1 and tested on batch $K$, and report the results in Table III. The performance variations of all methods are illustrated in Fig. 7. From Table III and Fig. 7, we have the following observations:

1. Manifold regularization based combined kernel (ML-comgfk) achieves an average accuracy of 79.6% and outperforms other SVM based machine learning algorithms and single kernel methods, which demonstrates that manifold learning and combined kernel can improve the classification accuracy, but limited capacity.
2. The classifier ensemble can improve the performance of the dataset with drift noise (an average accuracy of 80.0%). However, many base classifiers should be trained using the source data for ensemble, and it has no domain adaptability when tested on the data from target domains, which has been well referred in the proposed DAELM.
3. The proposed DAELM methods perform much better (91.82%) than all other existing methods for different tasks in recognition tested on drifted data. The robustness of the proposed methods with domain adaptability is proved for drift compensation in E-nose.

For studying the variations of recognition accuracy with the number $k$ of labeled samples in target domain, different number $k$ from the set of {5, 10, 15, 20, 25, 30, 35, 40, 45, 50} is explored by Algorithm 3 (SSA) and the proposed DAELM framework. Specifically, we present comparisons with different number of labeled samples selected from target domains. For fair comparison with ELM, the labeled target samples are feed into ELM together with the source training samples. The results for experimental **Setting 1** and **Setting 2** are shown in Fig. 8 and Fig. 9, respectively, from which, we have:

1. The traditional ELM has little obvious improvement with the increase of the labeled samples from target domains, which clearly demonstrates that ELM has no the capability of knowledge adaptation.
2. Both DAELM-S and DAELM-T have significant enhancement in classification accuracy with increasing labeled data from target domain. Note that in batch 2 and batch 10 shown in Fig. 8, our DAELM is comparative to ELM. The possible reason may be that little drift exist in batch 2 that leads to the small difference in classification task. While the data in batch 10 may be seriously noised by drift, the E-nose system may lose recognition ability only using batch 1 (Setting 1) for training. The proposed DAELM is still much better than ELM when tested on the seriously noised batch 10 in Setting 2 (Fig. 9).
3. DAELM-S has superior performance to DAELM-T when the number $k$ of labeled target samples used in knowledge adaptation is smaller, because DAELM-T does not consider the source data in classifier learning. Additionally, with the increase of the number $k$, DAELM-T has a comparative performance with DAELM-S which maybe a better choice when only a small number of labeled samples in target domain are available.

Throughout the paper, the proposed DAELM framework is to cope with sensor drift in the perspective of machine learning in decision level, but not intuitively calibrate the single sensor response because the drift rules are difficult to be captured by some linear or nonlinear regression method due to its nonlinear/chaotic dynamic behavior. This work is to construct a learning framework with better knowledge adaptability and generalization capability to drift noise existing in dataset.

## V. CONCLUSION AND FUTURE WORK

In this paper, the sensor drift problem in electronic nose is addressed by a new knowledge adaptation based machine learning approach. We have proposed a new framework, referred to as Domain Adaptation Extreme Learning Machine (DAELM), for fast knowledge transfer. Specifically, two algorithms, called DAELM-S and DAELM-T are proposed for drift compensation. The former learns a robust classifier based on the source domain by leveraging a limited number of labeled samples from target domain. The latter learns a classifier based on a limited number of labeled data in target domain by leveraging a pre-learned base classifier in source domain. From the angle of machine learning, the proposed methods provide new perspectives for exploring ELM theory, and also inherit the advantages of ELM including the feature mapping with randomly generated input weights and bias, the analytically determined solutions, and good generalization. Another important contribution, the key of this paper, is an effective measure using domain adaptation and ELM framework to cope with sensor drift in E-nose. Experiment on a long term sensor drift data set collected by E-nose clearly demonstrates the efficacy of our proposed framework. Additionally, the proposed framework can realize the recognition directly from the output (16) or (26) of algorithm without any cumbersome measure, which is completely different from SVM based methods that multi-class problem should be divided into multiple binary classification using "one against one" or "one against all" strategy and obtain the predicted label by voting mechanism. It is worth noting that the training time and testing time of proposed algorithms costs about several seconds and microseconds, respectively, due to the analytically determined solutions intuitively without iterations in learning process.

In the future, we will investigate on-line domain adaptation for drift compensation in E-nose from the viewpoint of incremental learning. It would be of interest to explore the nonlinear dynamic behavior of drift by constructing on-line dynamic classifiers with knowledge adaptation.

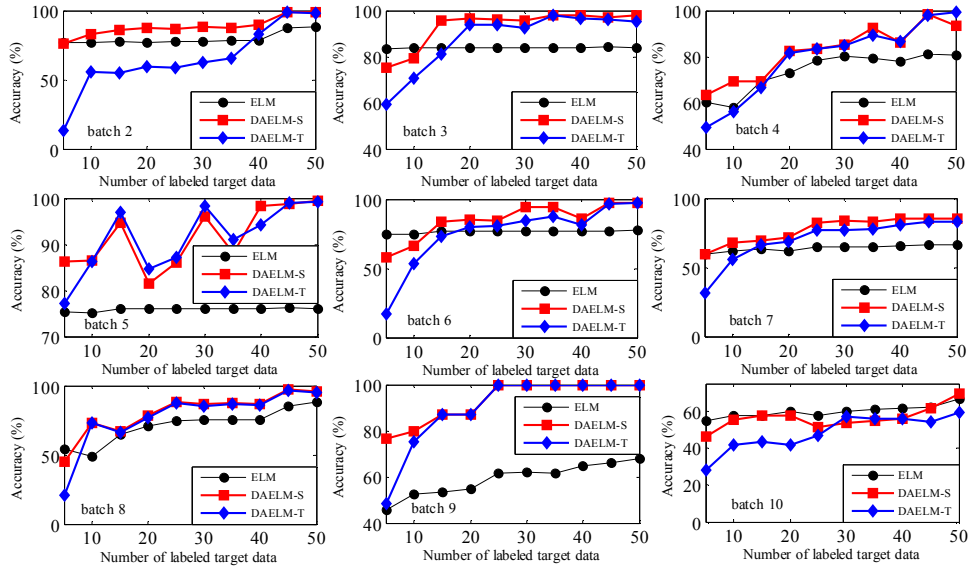

Fig. 8. Recognition accuracy under **Setting 1** with respect to different size of guide set (labeled samples from target domain)

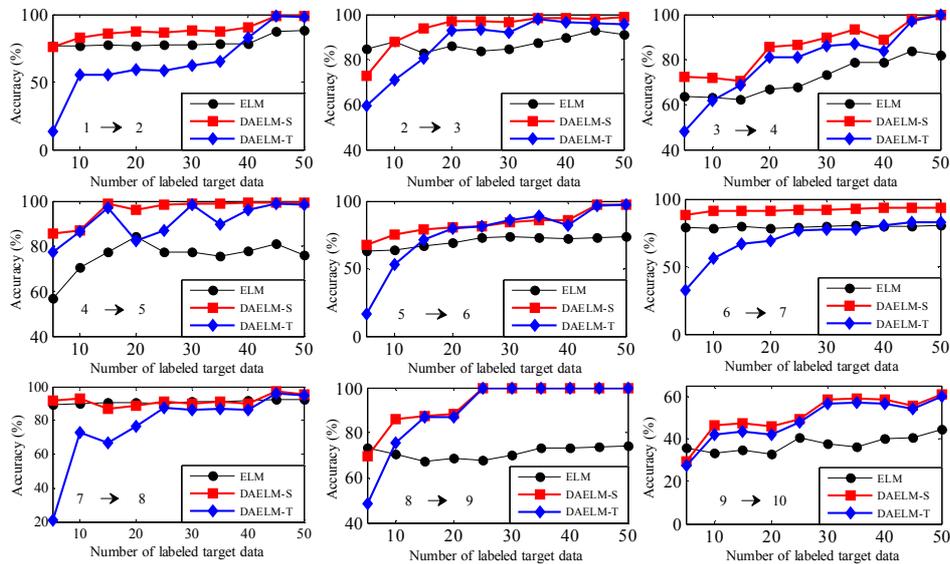

Fig. 9. Recognition accuracy under **Setting 2** with respect to different size of guide set (labeled samples from target domain)

ACKNOWLEDGMENT

The authors would like to thank Dr. Alexander Vergara in University of California San Diego for his provided sensor drift data in electronic nose. The authors would also like to express their sincere appreciation to the anonymous reviewers for their insightful comments.